\documentclass[conference]{IEEEtran}
\IEEEoverridecommandlockouts
\usepackage{cite}
\usepackage{pgfplots}
\usepackage{amsmath,amssymb,amsfonts}
\usepackage{booktabs}
\usepackage{algorithmic}
\usepackage{graphicx}
\usepackage{textcomp}
\usepackage{xcolor}
\usepackage{tikz}
\def\BibTeX{{\rm B\kern-.05em{\sc i\kern-.025em b}\kern-.08em
    T\kern-.1667em\lower.7ex\hbox{E}\kern-.125emX}}
\begin{document}

\title{Knowledge-guided EEG Representation Learning
}
\author{
  \IEEEauthorblockN{
    Aditya Kommineni\IEEEauthorrefmark{1},\quad
    Kleanthis Avramidis\IEEEauthorrefmark{1},\quad
    Richard Leahy\IEEEauthorrefmark{2},\quad
    Shrikanth Narayanan\IEEEauthorrefmark{1}\IEEEauthorrefmark{2}
  }
  \vspace{0.1cm}
  \IEEEauthorblockA{
    \IEEEauthorrefmark{1}Signal Analysis and Interpretation Lab,
    University of Southern California, Los Angeles, CA 90089
  }
  \IEEEauthorblockA{
    \IEEEauthorrefmark{2}Signal and Image Processing Institute,
    University of Southern California, Los Angeles, CA 90089
  }
}

\maketitle

\begin{abstract}
Self-supervised learning has produced impressive results in multimedia domains of audio, vision and speech. This paradigm is equally, if not more, relevant for the domain of biosignals, owing to the scarcity of labelled data in such scenarios. The ability to leverage large-scale unlabelled data to learn robust representations could help improve the performance of numerous inference tasks on biosignals. Given the inherent domain differences between multimedia modalities and biosignals, the established objectives for self-supervised learning may not translate well to this domain. Hence, there is an unmet need to adapt these methods to biosignal analysis. In this work we propose a self-supervised model for EEG, which provides robust performance and remarkable parameter efficiency by using state space-based deep learning architecture. We also propose a novel knowledge-guided pre-training objective that accounts for the idiosyncrasies of the EEG signal. The results indicate improved embedding representation learning and downstream performance compared to prior works on exemplary tasks. Also, the proposed objective significantly reduces the amount of pre-training data required to obtain performance equivalent to prior works.
\end{abstract}

\section{Introduction}
Self-supervised learning (SSL) has seen a growing interest in recent years, owing to its ability to leverage unlabelled data for learning effective representations. Domains such as text, vision and speech have been the primary beneficiaries of this paradigm enabling state of the art models for a wide range of tasks such as speech recognition, image and natural language understanding\cite{wav2vec, wav2vec2, gpt2, bert, image_vit}. Physiological signal and imaging modalities such as ECG, EEG, and fMRI can greatly benefit from such a pre-training setting. Considering that the availability of sufficiently labelled data in many scenarios is limited, and in most cases would require significant human expert annotations, it is desirable to create computational models that can learn in an unsupervised way, thereby enabling the model to transfer effectively to low-resource settings.
\par
There is limited work in biomedical settings that use large-scale data to directly learn meaningful signal representations. There have been some recent efforts to build such models using ECG\cite{wildecg}, fMRI\cite{fmri_ssl1} and EEG\cite{maeeg, bendr}. In the majority of these works, self-supervised objectives such as masked reconstruction, contrastive learning and augmentation-based objectives have been employed. These objectives have shown to provide robust results on modalities such as text, images and speech which typically have high signal-to-noise ratio (SNR). Therefore, masking and augmentation objectives assist the model in learning to decipher the semantic structure of the input, thereby learning meaningful representations. However, biomedical signals have unique characteristics; signals such as ECG and EEG have low SNR, their recording conditions can be heterogeneous, and the measuring devices can be diverse in terms of the number of channels, e.g., ranging anywhere from 1 to 12 for ECG and 2 to 128 for EEG. These factors pose challenges for a self-supervised model to learn from the aforementioned conventional approaches which further distort the signal in hand. Hence, there is a need for domain-aware self-supervised objectives that can better account for the nuances of biomedical signals.
\par

There have been previous attempts to build self-supervised models for EEG, especially in the context of sleep stage recognition. These works vary from employing contrastive predictive coding in \cite{sleep_ssl1}, contrastive learning through transformations of input signals such as addition of gaussian noise, time warping in \cite{sleep_ssl2} and a reconstruction-based objective using a masked autoencoder in \cite{maeeg}. Although these works report competitive results, the pre-training and fine-tuning data come from the same domain; 
the robustness of these SSL embeddings to diverse sleep EEG data or their ability to transfer to out of domain tasks are not well established. In a recent work, BENDR\cite{bendr}, the authors trained a generalized EEG model through pre-training on a large-scale EEG dataset. BENDR is based on transformer architecture with a convolution encoder. The embeddings from the encoder are masked, followed by a contrastive loss objective akin to wav2vec2\cite{wav2vec2}, an audio model. However, as noted earlier, there are differences between audio and EEG signals, including in terms of SNR patterns. Therefore, it is necessary to account for the characteristics of EEG while designing the training objectives to obtain robust performance.
\par
In addition to robust downstream performance, low parameter count is desired in most SSL models. However, prior EEG pre-training methods have used transformers\cite{vaswani2017attention} as their deep learning architecture, which are over-parameterized and are unable to encode long-range temporal dependencies. Recently, \textbf{Structured State Space for Sequence Modeling} (S4)\cite{s4_paper}, a State Space-based deep learning architecture was introduced showing impressive performance on time series tasks which makes them an ideal candidate for modelling EEG. Also, S4 layers are parameter and compute efficient compared to transformer models. In this work, we offer solutions to some of the drawbacks in prior works such as extremely large parameter count and lack of domain knowledge in pre-training objectives. The contributions of this work are threefold:
\begin{itemize}
    \item Showcase the superior performance of state space-based deep learning architectures (S4) compared to transformers in learning EEG representations. 
    \item Propose a novel knowledge-guided pre-training objective in order to better encode the characteristics of EEG in learned model embeddings. We demonstrate the effectiveness of the introduced loss on two downstream tasks.
    \item Demonstrate the robustness of proposed methods to data scarcity through empirical analysis by reducing the amount of pre-training and fine-tuning data.
\end{itemize}

\section{Datasets}
\subsection{Pre-training Dataset}
\begin{figure}
    \centering
    \includegraphics[width=0.4\textwidth]{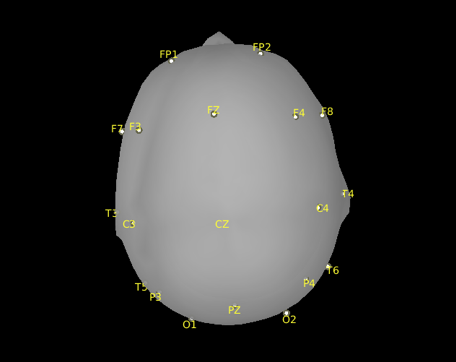}
    \caption{Montage of 19 EEG channels used for pre-training from TUEG corpus}
    \vspace{-0.2cm}
    \label{fig:montage}
\end{figure}

The Temple University EEG Corpus (TUEG)\cite{tuh_eeg_corpus} is used for pre-training. TUEG corpus is a large-scale clinical EEG recordings dataset from patients admitted to the Temple University Hospital. The recordings are diverse in terms of the demographics such as age, gender and various clinical conditions of participants. From the recordings, 19 channels [Fp1, F7, F3, Fz, F4, F8, Fp2, T3, C3, Cz, C4, T4, T5, P3, Pz, P4, T6, O1, O2] which are present in majority of the recordings are selected for pre-training. The montage of these channels are shown in Fig.\ref{fig:montage}. The selected subset of channels have representation from frontal, temporal and occipital regions of the brain. After limiting our data to these 19 channels, the total number of clinical recordings employed in the pre-training are 69,410 which amount to about 16,000 hours of raw EEG data.  
\par 
\begin{figure}
    \centering
    \includegraphics[scale=0.5]{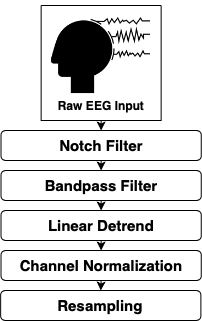}
    \caption{Pre-processing pipeline used for pre-training and fine-tuning datasets}
    \vspace{-0.2cm}
    \label{fig:preprocess}
\end{figure}

All recordings with sampling frequency 250 Hz or higher were used in pre-training. These recordings were resampled to 250 Hz. The pre-processing pipeline for the dataset is shown in Fig.\ref{fig:preprocess} and implemented using MNE-Python module \cite{mne_python}. First, a notch filter at 60 Hz is applied to remove power line noise. Following this, a bandpass filtering between [0.5, 50]Hz encompassing the spectrum of interest in EEG is performed. Then, linear detrending and channel-wise normalization are applied to account for heterogeneous recording environments. Finally, signals are resampled to the target 250 Hz frequency.\vspace{-0.2cm}
\subsection{Downstream Datasets}
The proposed pre-trained models are evaluated through performance on downstream tasks. The datasets used for downstream tasks include the MMI \cite{mmi_dataset} dataset and the BCI IV Task 2A dataset \cite{bci42a} obtained from Physionet\cite{goldberger2000physiobank}. The MMI dataset comprises 109 volunteers, out of which 4 subjects were ignored owing to unstable data recording. Each subject had performed binary tasks, i.e.,  opening/closing the right or left fist based on the direction in which the target appears on the screen. Each participant's recording consists about one to two minutes of 64 channel EEG data. The sampling rate is 160 Hz, requiring upsampling to 250 Hz, the sampling rate of pre-training data. In order to train the downstream models, the 19 channels considered for pre-training are selected and the remaining channels are discarded. During downstream evaluation, 5-fold cross validation is employed for all experiments.
\par 
BCI IV Task 2A is a motor imagery task consisting of EEG data from 9 subjects. This BCI task involved a cue-based imagined movement of the left arm, right arm, legs or tongue. The recording was conducted with 22 electrodes recorded at 250 Hz. However, there is a mismatch in the electrodes which the task uses, to the set of electrodes which were selected in from the TUEG corpus for pre-training. Hence, while training downstream models the channels are mapped based on spatial proximity to each of the 19 electrodes used in pre-training. Owing to this channel mismatch, the expected performance of pre-trained models for this task will be lower compared to conventional fully supervised models, which can use all the 22 electrodes and leverage the corresponding spatio-temporal correlations. For fine-tuning, since the dataset has only 9 subjects, leave one subject out cross validation (LOSOCV) is performed; the mean and standard deviations of accuracy for each experiment is reported. Both the datasets are balanced.

\section{Knowledge-Guided Objectives}
In this work, we investigate the potential of leveraging top-down knowledge about EEG in pre-training. Traditional SSL objectives such as masked reconstruction and contrastive learning involve distorting the input signal through addition of noise, masking, or reordering the time steps of input signals. Then, the model tries to reconstruct the input, encoding temporal structure and semantics during the training procedure. This objective works well in signals where the SNR is favorable. However, in the biomedical domain wherein the signals have inherently low SNR, using reconstruction as the only pre-training objective will lead to the model embedding characteristics of the noise in addition to the structure of the signal. One could alleviate this issue through grounding the SSL objective with less noisy features from the signal. 
\par
\begin{figure}
    \centering
    \includegraphics[width=0.4\textwidth]{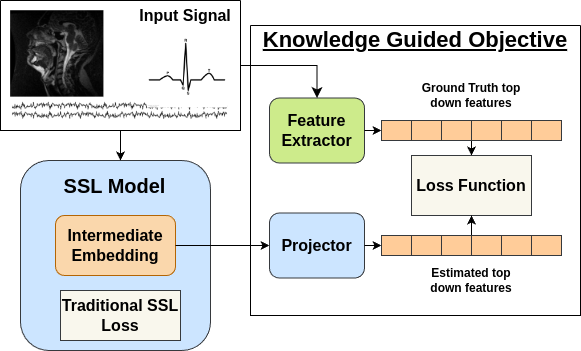}
    \caption{Illustration of our knowledge-guided objective which leverages top-down knowledge about signals, enabling learning of robust representations}
    \label{fig:KGO}
    \vspace{-0.4cm}
\end{figure}

 
Knowledge-guided objectives are defined as additional self-supervised objectives utilizing top-down signal processing features in order to enable the model to learn better representations. Fig.\ref{fig:KGO} shows the architecture for utilizing knowledge-guided objective along with traditional SSL objectives.
Raw signal ($\mathbf{x}_{\text{input}}$) is used as input to the model, followed by a traditional SSL objective loss.  Knowledge-driven objectives require the engineering of appropriate handcrafted features that encode the signal structure, i.e., one could choose connectivity matrices in fMRI, frequency band power in EEG or heart rate variability in ECG. Let $\mathbf{F}_{\text{ground truth}}$ be those features, which are directly computed from the input. Then, intermediate embeddings from the SSL model can be projected, i.e., through fully connected layers, to obtain $\mathbf{F}_{\text{estimated}}$. Eq.\eqref{kgl_loss} defines the knowledge-guided objective $\mathbf{L}_{\text{knowledge}}$ as the reconstruction loss of $\mathbf{F}_{\text{ground truth}}$ given $\mathbf{F}_{\text{estimated}}$. Appropriate loss function $\mathbf{g}$ is selected depending on the selected signal processing feature:
\begin{equation}\label{kgl_loss}
    \mathbf{L}_{\text{knowledge}} = \mathbf{g}(\mathbf{F}_{\text{ground truth}}, \mathbf{F}_{\text{estimated}}).
\end{equation}

\section{Model Architecture}
\begin{figure*}
    \centering
    \includegraphics[width=\textwidth]{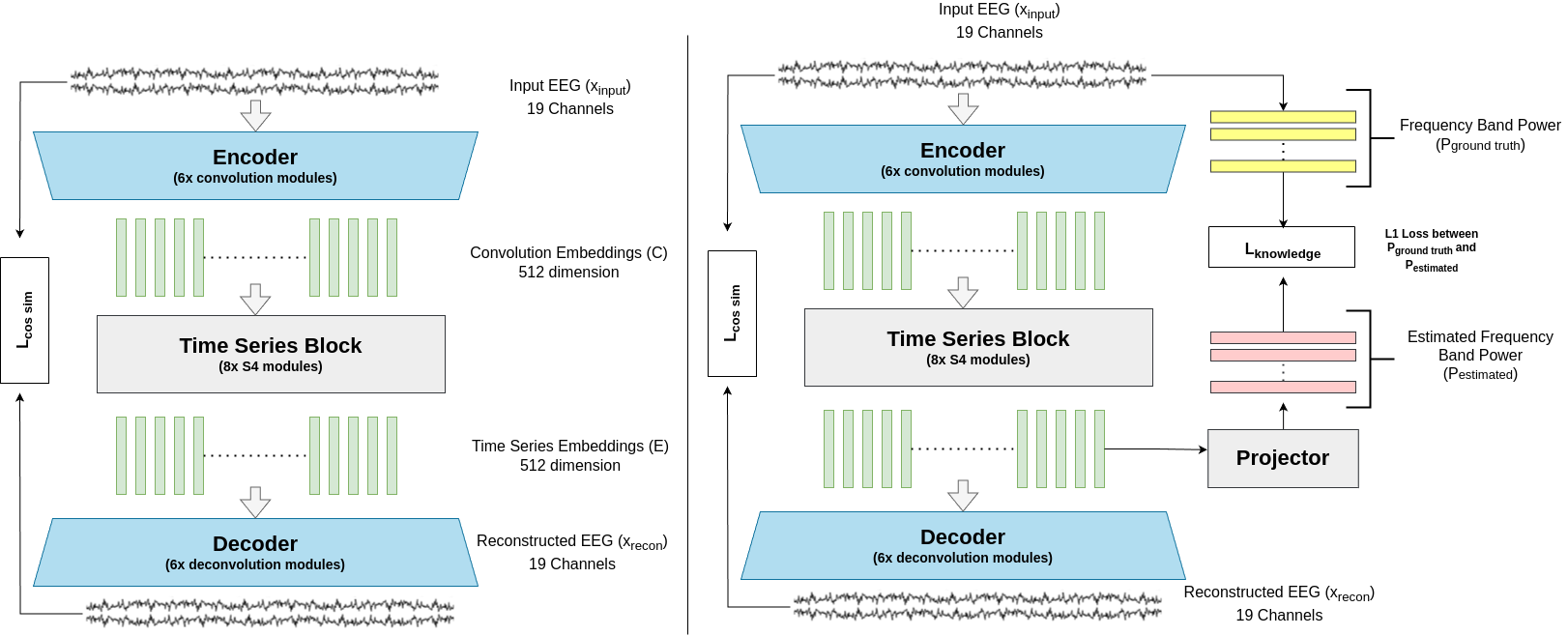}
    \caption{ Comparison between model architecture and SSL objectives for proposed pre-trained models. (\textbf{Left}) Vanilla S4 is trained on reconstruction loss. (\textbf{Right}) Knowledge-guided S4 employs an L1 loss between actual and estimated frequency band power in addition to reconstruction loss.}
    \label{fig:ssl_arch}
\end{figure*}

Two SSL models are proposed based on the objectives described in the previous section, namely \textit{Vanilla S4} and \textit{Knowledge-guided S4}. Both models use identical backbone architectures as seen in Fig.~\ref{fig:ssl_arch} but vanilla S4 is pre-trained on traditional SSL objectives alone whereas knowledge-guided S4 is trained on a combination of traditional SSL objectives and knowledge-guided objectives.

\par 
Fig.\ref{fig:ssl_arch} shows the backbone architecture, an encoder-decoder framework with a time-series connecting block. The encoder acts as a bottleneck, wherein the intermediate embeddings aim to learn effective temporal representations according to the defined self-supervised objectives. The encoder is composed of 6 convolution modules, each being a combination of a 1D convolution layer, a dropout ($p=0.3$) layer, layer normalization and GELU activation. Convolution layers have been shown to act as learnable filters for SSL pre-training in domains such as speech~\cite{wav2vec2}. The encoder takes chunks of 60s pre-processed EEG signals $(\mathbf{x}_{\text{input}}\in\mathbb{R}^{{\text{N}}_{\text{channels}} \times \text{N}_{\text{time steps}}})$ as inputs where $\text{N}_{\text{channels}}$ is the number of EEG channels (in this case 19) and $N_{\text{time steps}}$  is the time steps equals 15360 and produces convolution embeddings $(\mathbf{C} \in \mathbb{R}^{\text{N}_{\text{dim}} \times \text{N}_{\text{embeddings}}})$ where size of embeddings $\text{N}_{\text{dim}}$ is 512 and number of embeddings per input $\text{N}_{\text{embeddings}}$ is 240. The encoded embeddings effectively represent a window of about 250 ms, which ensures that the model is able to capture temporal dynamics suitable for a range of tasks with both short and long-range temporal dependencies. 
\par
The time-series block is composed of a linear layer followed by eight S4 modules. Each module is composed of a S4 kernel estimator, Gated Linear Unit (GLU) activation, dropout ($p=0.3$) and layer normalization. Time series block takes $\mathbf{C}$ as inputs and produces time series embeddings $(\mathbf{E} \in \mathbb{R}^{\text{N}_{\text{dim}} \times \text{N}_{\text{embeddings}}})$ at the output. Since reconstruction objectives are utilized, in addition to encoder and time series block, a decoder is employed to project the time series embeddings back to the dimensions of input signal, resulting in the reconstructed signal $(\mathbf{x}_{\text{recon}}\in \mathbb{R}^{\text{N}_{\text{channels}} \times \text{N}_{\text{time steps}}})$. The decoder is composed of 6 deconvolution modules, each module contains a 1D transpose convolution layer along with a dropout ($p=0.3$) layer, layer normalization and GELU non linearity.
\subsection{Vanilla S4}
The architecture for vanilla S4 is as seen in Fig.\ref{fig:ssl_arch} (left). This model is pre-trained using traditional SSL objectives. Previous works MAEEG and BENDR are trained using reconstruction and contrastive-based SSL objectives, respectively. Since they are trained on identical architectures, the performance of these objectives can be compared. MAEEG outperforms BENDR when fine-tuned for downstream tasks. This suggests that reconstruction objectives could be better for modelling EEG signals. Hence, reconstruction objective based on cosine similarity [$\frac{\mathbf{a} \cdot \mathbf{b}}{\|\mathbf{a}\| \cdot \|\mathbf{b}\|}$] is used in vanilla S4. The loss is calculated according to Eq.\eqref{cosine_loss}, $\mathbf{x}_{\text{input}}^{i}$ and $\mathbf{x}_{\text{recon}}^{i}$ are the $i^{th}$ channel in input and reconstructed EEG signal, respectively:
\begin{equation}\label{cosine_loss}
L_{\text{cos sim}} = 1 - \frac{1}{\text{N}_\text{channels}} \sum_{i=1}^{\text{N}_\text{channels}} \frac{(\mathbf{x}_{\text{input}}^{i})^{\text{T}} (\mathbf{x}_{\text{recon}}^{i})}{\|\mathbf{x}_{\text{input}}^{i}\| \|\mathbf{x}_{\text{recon}}^{i}\|}.
\end{equation}

\subsection{Knowledge-guided S4}
This model employs a combination of previously employed reconstruction objective and knowledge-guided objective. Frequency band power is selected as the signal processing feature to provide the knowledge guidance for the model during SSL training. Delta [0.5-4 Hz], theta [4-8 Hz], alpha [8-13 Hz], beta [14-30 Hz] and gamma [$>$30 Hz] bands are included in feature computation. These features have been used to explain various functions of the brain such as sleep, motor movements and cognitive processing\cite{saby2012utility, newson2019eeg}. Hence, explicit knowledge about these features would help the model learn better time series embeddings ($\mathbf{E}$). For accurately computing the power of delta band, which has frequencies as small as 0.5 Hz, the time window of Power Spectral Density (PSD) needs to be at least one time period (2s). Hence, ground truth frequency band power $\mathbf{P_{\text{ground truth}}}$ is estimated with a non overlapping 4 second time window. As mentioned previously, each embedding of $\mathbf{E}$ effectively encodes information from 250 ms of the input EEG time series. Hence, in order to estimate the frequency band power from the time series embeddings, mean of 16 embeddings is taken which is equivalent to 4 seconds in the time domain. Then, the resulting embeddings are passed into the projector as in Fig.\ref{fig:ssl_arch} (right). The projector is a single fully connected layer which projects the embeddings from dimension ($\text{N}_{\text{dim}} = 512$) to frequency band power resulting in estimated frequency power bands $\mathbf{P}_{\text{estimated}}$. 

\par
Knowledge-guided loss is defined as the L1 loss between $\mathbf{P}_{\text{ground truth}}$ and $\mathbf{P}_{\text{estimated}}$ as in Eq.~\eqref{loss_knowledge} wherein $\text{N}_{\text{channels}}$, $\text{N}_{\text{bands}}$ and $\text{N}_{\text{time steps}}$ are the number of channels in the signal, number of frequency bands considered and the number of time steps for which power is computed respectively:
\begin{equation}\label{loss_knowledge}
L_{\text{knowledge}} = \sum_{i=1}^{\text{N}_{\text{channels}}} \sum_{j=1}^{\text{N}_{\text{bands}}} \sum_{k=1}^{\text{N}_{\text{time steps}}} \left| \mathbf{P}_{\text{ground truth}}^{(i, j, k)} - \mathbf{P}_{\text{estimated}}^{(i, j, k)} \right|,
\end{equation}
\begin{equation}\label{total}
    L_{\text{combined}} = L_{\text{cos sim}} + \lambda L_{\text{knowledge}}.
\end{equation}
For pre-training, a combination of $L_{\text{cos sim}}$ and $L_{\text{knowledge}}$ as in Eq.\eqref{total} is used. Here, $\lambda$ is a hyperparameter whose value is set to 5. In this manner, the model learns temporal dependencies in the signal and at the same time, these learned relations reflect the signal knowledge in terms of frequency band power.

\section{Experiments}

The previous section provides the implementation details for the proposed pre-training models, vanilla S4 and knowledge-guided S4. Following the pre-training, these models are evaluated through fine-tuning on downstream tasks. MMI and BCI IV 2A are the downstream datasets used for fine-tuning. During fine-tuning, the decoder for both the pre-trained models is discarded. In all experiments, the encoder of pre-trained models is frozen. The time series embeddings ($\mathbf{E}$) are averaged to obtain a single 512 dimensional embedding which is used as an input to a classification head. It is composed of fully connected layers projecting the embedding to number of classes in the task. 
\par 
For each pre-trained model, three fine-tuned settings are investigated. Table~\ref{main_results} shows the results for these three settings, namely Last S4 module, All S4 modules and Linear Probe. 
\begin{itemize}
    \item \textbf{Last S4 module} refers to the last S4 module in the time series block being trainable during fine-tuning.
    \item \textbf{All S4 modules} refers to the entire time series block being trainable, i.e., eight S4 modules.
    \item \textbf{Linear Probe} refers to the setting wherein pre-trained models are used akin to traditional signal processing features. Both the encoder and time series block are frozen during fine-tuning and classification head is composed of fully connected layers. $N_{fc}$ denotes the number of fully connected layers in the classification head.
\end{itemize}
\par

\textbf{Fully trainable model} does not have a self supervised objective, i.e., it is not pre-trained. The weights of the encoder and S4 layers are randomly initialized. For this model, during the fine-tuning stage all the parameters in the encoder and S4 layers are unfrozen. This model will act as a baseline for the performance of S4 models without pre-training. The best model for each downstream dataset, as mentioned in BENDR, will be used as a baseline for transformer models. Additionally, the performance of pre-trained models is compared to some fully supervised works. These models use all the channels present in the data thereby aided by the resulting spatio-temporal correlations which the pre-trained models cannot.
\par
Pre-training is performed for 500k iterations using a batch size of 32 and Adam optimizer with a learning rate $0.001$. For fine-tuning, a batch size of 64 is used for both the downstream tasks. Learning rate is tuned within $0.001, 0.0001$.

\section{Results}
\subsection{Downstream Performance}
Table~\ref{main_results} shows the results for the various configurations of vanilla S4 and knowledge-guided S4 and compares them to relevant baselines. For all datasets, accuracy has been chosen as the evaluation metric with the mean and standard deviation reported for each models performance.
\par
Firstly, knowledge-guided S4 outperforms all the other pre-trained baselines and vanilla S4 for both MMI and BCI IV 2A datasets. In the case of BCI IV 2A, knowledge-guided S4 outperforms BENDR by 12.65\% relatively. Additionally, knowledge-guided S4 shows improvement in performance when compared to vanilla S4 on both tasks, validating the hypothesis that knowledge-guided objectives aid in providing a better convergence point for the pre-trained models.
\par
When compared to fully supervised baselines such as \cite{adaptivecnn, eegnet_ensemble}, the proposed knowledge-guided S4 model outperforms in the case of MMI dataset. Additionally, it must be noted that these supervised baselines use all the 64 channels present in the data. However, the proposed models are constrained to use the 19 channels which were chosen in pre-training. In the case of BCI IV 2A dataset, the performance is better compared to fully supervised baseline MCNN\cite{mcnn}, but fails to outperform CCNN\cite{mcnn}. It may arise from the channel mismatch between the electrodes used in pre-training and the ones in the BCI IV 2A dataset, resulting in the electrodes being mapped based on spatial proximity. This results in suboptimal performance owing to the model unable to adapt to the new set of channels.
\par
The performance of the fully trainable model, i.e., the model without pre-training, is comparable to that of BENDR which is a pre-trained transformer model. This showcases superior performance of S4 layers in modelling time series data, especially biomedical signals compared to transformers. S4 models are also light in terms of the number of parameters in the model. In the proposed models, i.e., vanilla S4 and knowledge-guided S4, the total number of parameters in the models are about 13M, whereas the transformer-based BENDR has about one billion parameters.
\par 
In order to gauge the effectiveness of the learnt time series embeddings during pre-training, linear probing is performed, i.e., encoder and time series block are frozen and a classification head is added with fully connected layers. Both vanilla S4 and knowledge-guided S4 show impressive accuracies of over 73\% with just a single fully connected layer on the MMI dataset demonstrating the discriminating ability of the embeddings. For the BCI IV 2A dataset, although the fully connected layer models perform much better than chance, but underperform compared to baselines. This is due to the previously mentioned channel mismatch. When the last S4 module is made trainable, the performance improves significantly to outperform BENDR and fully trainable model.

\begin{table}
    \centering
    \caption{Evaluation Accuracy (\%) of the pre-trained models on downstream datasets. 5-fold cross validation is used for MMI and leave one subject out cross validation for BCI IV 2A}
    \vspace{-0.2cm}
    \begin{tabular}{lcc}
        \toprule
        \textbf{Dataset} & \textbf{MMI}  & \textbf{BCI IV 2A} \\
        \midrule
        Random & 50 & 25 \\
        \midrule
        \textbf{Vanilla S4} (Ours) & & \\
        \quad Linear Probe $(N_{fc}=1)$ & 75.52 $\pm$ 3.65 & 32.58 $\pm$ 3.2 \\
        \quad Linear Probe $(N_{fc}=1)$ & 81.63 $\pm$ 2.22 & 33.83 $\pm$ 2.52 \\
        \quad Last S4 layer & 83.97 $\pm$ 1.55 & 44.42 $\pm$ 4.91 \\
        \quad All S4 layers & 84.35 $\pm$ 1.42 & 46.81 $\pm$ 3.96 \\
        \midrule
        \textbf{Knowledge-guided S4} (Ours) & & \\
        \quad Linear Probe $(N_{fc}=1)$ & 73.5 $\pm$ 3.18 & 31.59 $\pm$ 3.35 \\
        \quad Linear Probe $(N_{fc}=2$) & 80.52 $\pm$ 2.51 & 34.56 $\pm$ 3.49 \\
        \quad Last S4 layer & 85.03 $\pm$ 1.20 & 47.29 $\pm$ 6.2 \\
        \quad All S4 layers & \textbf{87.12} $\pm$ 1.15 & \textbf{47.99} $\pm$ 5.70 \\
        \midrule
        \textbf{Pre training Baselines} & & \\
        \quad BENDR\cite{bendr} & 86.7 & 42.6 \\
        \midrule
        \textbf{Fully Supervised Baselines} & & \\
        \quad Fully Trainable (Ours) & 86.6 $\pm$ 1.37 & 42.8 $\pm$ 3.4 \\
        \quad EEGnet based ensemble\cite{eegnet_ensemble} & 86.36 & - \\
        \quad Deep Subject Adaptive CNN\cite{adaptivecnn} & \textbf{86.93} & - \\
        \quad MCNN\cite{mcnn} & - & 42.1 \\
        \quad CCNN\cite{mcnn} & - & \textbf{55.3} \\
        \bottomrule
    \end{tabular}
    \vspace{-0.5cm}
    \label{main_results}
\end{table}

\subsection{Data Efficiency}
Data efficiency is analyzed along two dimensions, i.e., pre-training and fine-tuning data by looking into the relationship between data size and corresponding performance boost.
\subsubsection{Pre-training Data}
The amount of data used for pre-training is an important factor while employing self-supervised learning paradigm. The ability to use lesser data without negatively affecting the transferability of the model is of great interest. Here, knowledge-guided S4 models are pre-trained with varying proportions (50\%, 10\%, 1\% and 0.1\%) of the TUEG corpus and evaluated on downstream tasks. The subsets of pre-training dataset of different proportions are created through sub sampling the original TUEG corpus.
\par
Downstream performance of knowledge-guided S4 models at different percentages of pre-training data for the all S4 module setting are as given in Table~\ref{tab:pre-training_data_scarcity}. Validation procedure mentioned previously has been used for both datasets. From the results, it is observed that despite using as low as 1\% of the TUEG corpus, there is negligible performance drop across both the downstream datasets. This implies diminishing returns with increase in pre-training data size, emphasizing the importance of employing signal appropriate pre-training objectives. When compared to BENDR and vanilla S4 models, knowledge-guided S4 models outperform despite using 50, 10 and 1\% of the TUEG corpus portraying the effectiveness of the proposed knowledge-guided objective as opposed to employing reconstruction loss alone. 

\subsubsection{Fine-tuning Data}
\begin{figure}
    \begin{tikzpicture}
        \begin{axis}[
            xlabel={Percentage of fine-tuning data },
            ylabel={Accuracy (\%)},
            xtick=data,
            symbolic x coords={10, 30, 50, 100},
            x tick label style={rotate=45,anchor=east},
            log basis x={10},
            legend pos=north west,
            grid=major,
            grid style={dashed,gray!30},
            legend entries={Knowledge-guided S4, Vanilla S4, Fully Trainable},
            width=0.45\textwidth,
            height=7cm,
            nodes near coords,
            nodes near coords align={vertical},
            error bars/y dir=both,
            error bars/y explicit,
        ]
            \addplot[mark=*, blue] coordinates {
                (10,33.68)
                (30,38.5)
                (50,41.68)
                (100,47.99)
            };
            \addplot[mark=square*, red] coordinates {
                (10,34.18)
                (30,36.68) 
                (50,39.64)
                (100,46.81) 
            };
            \addplot[mark=triangle*, green] coordinates {
                (10,31.79)
                (30,33.27)
                (50,37.82) 
                (100,42.8)
            };
        \end{axis}
    \end{tikzpicture}
    \vspace{-0.4cm}
    \caption{Classification performance on BCI IV 2A dataset with decreasing fine-tuning data percentage. Results are illustrated for 100, 50, 30 and 10\%.}
    \label{fig:fine-tuning_graph_errorbars}
    \vspace{-0.5cm}
\end{figure}
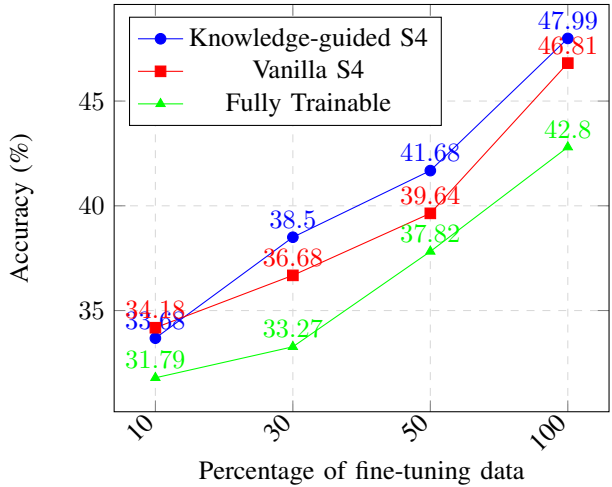

As mentioned previously, the amount of labelled data which can be obtained in a majority of EEG applications is limited. We evaluate the performance of the models in such data scarce settings through testing on a reduced proportion of the available downstream training samples for the BCI IV 2A dataset in specific. From Fig.~\ref{fig:fine-tuning_graph_errorbars}, all three models, i.e., fully trainable, vanilla S4 and knowledge-guided S4 perform significantly better than chance (25\%) for the 4-class classification problem, reflecting the ability of S4 models to learn in data scarce scenarios.

\begin{table}
    \centering
    \caption{Downstream evaluation accuracy (\%) of knowledge-guided S4 model on varying percentages of the original TUEG corpus.}
    \begin{tabular}{cccc}
        \toprule
        \textbf{Percentage of pre-training data (\%)} & \textbf{MMI} & \textbf{BCI IV 2A} \\
        \midrule
        0.10 & 85.99 $\pm$ 1.20 & 45.83 $\pm$ 4.47 \\
        1 & 86.34 $\pm$ 1.49 & 47.78 $\pm$ 5.03 \\
        10 & 87.11 $\pm$ 1.61 & 48.22 $\pm$ 6.2 \\
        50 & 86.77 $\pm$ 1.42 & 49.11 $\pm$ 6.1 \\
        100 & 87.12 $\pm$ 1.15 & 47.99 $\pm$ 5.70\\
        \bottomrule
    \end{tabular}
    \label{tab:pre-training_data_scarcity}
\end{table}

\section{Conclusion}
This work introduces a novel knowledge-guided objective for self-supervised learning which leverages top-down knowledge about the signals and provides simpler objectives compared to traditional self-supervised approaches such as masking and reconstruction. The effectiveness of the proposed knowledge-guided objective is demonstrated empirically through improved performance on two downstream tasks (motor movement, and motor imagery). Additionally, superior data efficiency of the proposed objective--both in terms of the pre-training and the fine-tuning data--is demonstrated, which is of importance in the biomedical domain. Future work will further investigate the ability of the models to translate well across different channel configurations. One promising approach could be employing the source space, which would homogenize the signal unlike the various montages in the sensor space. Additionally, we plan to extend these objectives to other biomedical signals with similar channel dependencies.

\bibliographystyle{unsrt}
\bibliography{refs}

\end{document}